\def\year{2018}\relax
\documentclass[letterpaper]{article} %DO NOT CHANGE THIS
\usepackage{aaai18}  %Required
\usepackage{times}  %Required
\usepackage{helvet}  %Required
\usepackage{courier}  %Required
\usepackage{url}  %Required
\usepackage{graphicx}  %Required
\usepackage{amsmath}
\usepackage{CJKutf8} % to input Chinese characters
\usepackage{subfigure}
\usepackage{cases}
\usepackage{enumerate}
\usepackage{graphicx}
\usepackage{diagbox}
\usepackage{makebox}
\usepackage{multirow}
\usepackage{bm}
\usepackage{amsfonts}
\usepackage{array}
\usepackage{flushend}
\newcolumntype{P}[1]{>{\centering\arraybackslash}p{#1}} % to center tabular value
\begin{document}
\title{Flexible End-to-End Dialogue System for Knowledge Grounded Conversation}
\author{ Wenya Zhu \textsuperscript{1} , Kaixiang Mo\textsuperscript{1}, 
               Yu Zhang\textsuperscript{1},  Zhangbin Zhu\textsuperscript{2}, 
              Xuezheng Peng \textsuperscript{2}, Qiang Yang\textsuperscript{1}\\
		\textsuperscript{1}{Hong Kong University of Science and Technology, Hong Kong, China}\\
		\textsuperscript{2}{Mobile Internet Group, Tecent Inc., Shenzhen, China}\\
}

\maketitle
\begin{abstract}
In knowledge grounded conversation, domain knowledge plays an important role in a special domain such as Music. The response of knowledge grounded conversation might contain multiple answer entities or no entity at all. Although existing generative question answering (QA) systems can be applied to knowledge grounded conversation, they either have at most one entity in a response or cannot deal with out-of-vocabulary entities. We propose a fully data-driven generative dialogue system GenDS that is capable of generating responses based on input message and related knowledge base (KB). To generate arbitrary number of answer entities even when these entities never appear in the training set, we design a dynamic knowledge enquirer which selects different answer entities at different positions in a single response, according to different local context. It does not rely on the representations of entities, enabling our model deal with out-of-vocabulary entities. We collect a human-human conversation data (ConversMusic) with knowledge annotations. The proposed method is evaluated on CoversMusic and a public question answering dataset. Our proposed GenDS system outperforms baseline methods significantly in terms of the BLEU, entity accuracy, entity recall and human evaluation. Moreover,the experiments also demonstrate that GenDS works better even on small datasets.

\end{abstract}

\section{Introduction}
% Neural conversation model
%Non-goal-driven Dialogue Systems devotes to generate fluent and relevant responses given the conversation history. Many researchers focus on fully data-driven dialogue system, which learn neural conversation model directly from large copora \cite{serban2015survey}. For example, the Seq2seq \cite{sutskever2014sequence} model, which can predict the target sequence given the source sequence, is the most popular neural model for data-driven dialogue system\cite{serban2017multiresolution}\cite{sordoni2015neural}. However, such systems tend to generate safe, commonplace, unrelavant and uninformative response (e.g., I don’t know) \cite{li2015diversity}\cite{serban2017multiresolution}. %That is to say,  they cannot responding substantively. We put this down to the disconnecting of external knowledge

\begin{figure}[!htbp] 
\centering\includegraphics[width=3.5in]{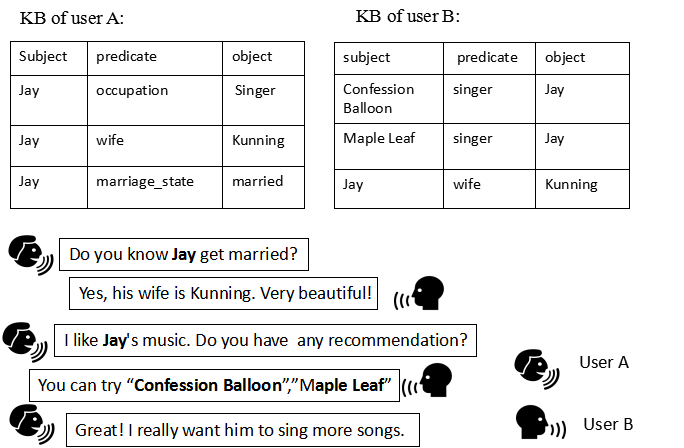} 
%\centering\includegraphics[width=3in]{aaai_18_fig_1_dialogue_sample.png} 
\caption{An example of knowledge-grounded non-goal-driven dialogue between two user A and B. Each user has its own private KB. In this example, two users are talking about the singer Jay Chou.}\label{Dialogue example} 
\end{figure}

% Knowledge is important for conversation.
Daily conversations generally depends on individual's knowledge. This is known as knowledge grounded conversation \cite{han2015exploiting,ghazvininejad2017knowledge}. In Figure~\ref{Dialogue example},  we show an example of knowledge grounded conversation, in which two friends are talking about music on their own knowledge base. To reply ``I like Jay's music. Do you have any recommendation?", user A has to know some songs of the singer. %We name such conversation type as knowledge-grounded non-goal-driven dialogues system.
% Example of knowledge grounded conversation.
% Knoledge grounded conversation is different from QA
It is necessary to emphasize that the knowledge grounded conversation is different from QA \cite{yin2015neural,he2017hegenerating}, as the former does not limit the number of entities in responses. For example, in Figure~\ref{Dialogue example}, two friends are talking about a singer named Jay, user A does not need any knowledge when replying ``Great! I hope he can sing more songs``, and the entities to reply the question "I like Jay's music. Do you have any recommendation?"  are also not unique. We can regard QA as a special case of knowledge grounded conversation. 

% Existing related work.
Han et al. \cite{han2015exploiting} first built rule-based chit-chat dialogue systems with structured knowledge. Ghazvininejad et al.\cite{ghazvininejad2017knowledge} took unstructured text as external knowledge to enhance traditional chit-chat dialogue systems by replying more informatively. In ~\cite{ghazvininejad2017knowledge}, all entities are presented by distributed representations, thus a large amount of data are required to figure out the relations between entities. In this paper, we aim to build an end-to-end knowledge grounded conversation model with structured KB, which is more effective to represent the relations between entities. 

Another category of related work is generative QA with KB ~\cite{haoend2017QA,yin2015neural,he2017hegenerating}. Many existing end-to-end QA models, such as GenQA~\cite{yin2015neural} and COREQA~\cite{he2017hegenerating}, can generate responses with facts retrieved from KB. However, these models are not able to deal with out-of-vocabulary entities. GenQA~\cite{yin2015neural} even cannot generate multiple entities. Besides, their decoding process relied on the representations of entities learned from conversation data, where the entities are sparse in the conversations (see details in Dataset section). Hence, their models require a large amount of data.

% The proposed method.
In this paper, we propose a fully data-driven generative dialogue system called GenDS, which can generate responses based on input message and structured KB. We introduce a dynamic knowledge enquirer, in order to generate an arbitrary number of  entities (even when the entities never appear in the training set). Based on local contexts, the system can select  entities at various positions within a single response. Specifically,  the dynamic knowledge enquirer updates the entity generation probability based on the previous context. This is independent of the representations of entities, which enables our model to handle out-of-vocabulary entities. With experiments, we find that the dynamic knowledge enquirer can punish repeated entities and remember the co-occurred entities. 

% The proposed method.
%In this paper, we first use the structured KB as the external knowledge and build the end-to-end non-goal-oriented dialogue system. We take advantage of the neural model on modeling the flexibility and diversity of language, and the structured KB on connecting entities with relations. The relevant facts are obtained according to the entities mentioned in the question. The specifically designed decoder and relevant facts are used as the target word source. Another neural network can select the target word source at each decoder time stamp. 
%To let the relevant but not exact reply acceptable, we train our system with multi-task learning \cite{luong2015multi}\cite{zhang2017survey}, where the response of one task is the same as the ground truth, and that of the other task is the ground truth but replacing the entities with their type. To facilitate the research on knowledge-grounded non-goal-oriented dialogue system, we create a new human-human conversation data. We conduct experiments both on our collected conversation and an public QA data.

% Contribution
In summary, our contributions are three-fold. 
\begin{itemize}
\item We propose a fully data-driven generative dialogue system GenDS that is capable of generating responses with any number of entities. A dynamic knowledge enquirer is proposed to select different answer entities according to different local contexts.
\item We have collected a real world dataset named ConversMusic with human annotation, which will be released to the public after acceptance. To the best of our knowledge, there is no public conversation dataset with annotated knowledge.
\item We evaluate our method on two datasets, namely a collected real-world music chatting dataset (ConversMusic) and a public question answering dataset. We show that the proposed method improves baseline models in terms of BLEU, entity accuracy, entity recall and human evaluation. 
\end{itemize}

\section{Related Work}
\textbf{Data-driven non-goal-oriented dialogue system} Recently, there is a trend towards developing fully data-driven dialogue systems. Seq2Seq \cite{sutskever2014sequence} learning, which can predict target sequence given source sequence, has been widely applied in such systems. Specifically, Shang et al. \cite{shang2015neural} first utilized the encoder and decoder framework to generate responses on micro-blogging websites. Sordoni et al. \cite{sordoni2015neural} extended it by conditioning the response generation on context vector, which is the encoding vector of three past consecutive utterances. Yao er al. \cite{yao2015attention} employed an intention network to maintain the relevance of responses. Serban et al. \cite{serban2016building} built an end-to-end dialogue system with generative hierarchical neural network. Serban et al. \cite{serban2017hierarchical} designed a latent variable RNN to model the complex dependencies between the sub-sequences, where the latent variables represent semantics of the sentence. \\
\textbf{Data-driven non-goal-oriented dialogue system with external knowledge}  Recent studies realized that non-goal-driven dialogue systems cannot reply substantively. This is caused by the isolation from external knowledge. Therefore, researchers began to incorporate external knowledge to enhance reply generation. Han et al. \cite{han2015exploiting} proposed a rule-based dialogue system by filling the response templates with retrieved KB. Ghazvininejad et al.\cite{ghazvininejad2017knowledge} utilized external textual information as the unstructured knowledge. As demonstrated, the external textual information can convey more relevant information to responses. \\
\textbf{Data-driven QA with external knowledge} Some recent work used external structured knowledge graph to build end-to-end question answering systems. Yin et al. \cite{yin2015neural} proposed a seq2seq-based model where answers were generated in two ways, where one was based on a language model and the other was by some entities retrieved from the KB. He et al. \cite{he2017hegenerating} further introduced another generation mechanism: copying words from original question. Besides, they also studied the cases where questions require multiple facts.

\section{Problem}
In this section, we first define notations and then introduce the problem setting.

\subsection{Notation}
Matrices are denoted in bold capital case, column vectors are in  bold lower case and scalars are in lower case. ${DNN}_i$ denotes $i$-layer neural network function. An input message is denoted by $\bm{X} = \{x_1,x_2,\cdots,x_{M_T}\}$ where $M_T$ is the number of words in the question. A response is denoted by $\bm{Y} = \{y_1,y_2,\cdots,y_{R_T}\}$ where ${R_T}$ is the number of words in the response. A knowledge fact represented as a triple (\textit{subject},\textit{predicate},\textit{object}), denotes as $\tau = \{\tau_{s},\tau_{p},\tau_{o}\}$. Specifically, subjects and objects are also known as entities, and predicate is the relation between the subject and object. A knowledge base is a set of all possible facts, denoted by $\mathcal{K} = \{(\tau_{si},\tau_{pi},\tau_{oi})\}_{i=1}^{K_N}$, where $K_N$ is the number of facts in the knowledge base.

\subsection{Problem Definition}
Given an input message, the problem is to generate an appropriate response based on knowledge base. The system firstly retrieves an arbitrary number of related facts from the knowledge base, then generates a response with the relevant facts. All the input messages and responses are comprised of two kinds of words, respectively the common words and the knowledge words. The knowledge words are entities in the knowledge base \footnote{We use knowledge words and entities interchangeably.}, while the rest are common words. 
The inputs of the problem are:
\begin{enumerate}
\item An input message $\bm{X}$. 
\item A knowledge base $\mathcal{K}$ containing all possible facts. 
\item A list of entity types $\mathcal{T}$. 
\end{enumerate}
The output of the problem is:
\begin{enumerate}
\item A response $\bm{Y}$. The response might contain arbitrary number of common and knowledge words.
\end{enumerate}
%What's in the training set?
For model training, the related facts $\tau$ for each message response pair are provided as training data, denoted by $\mathcal{D} = \{(\bm{X}_i,\bm{Y}_i,\{\tau\})\}_{i=1}^{D_N}$, where $\{\tau\}$ are the facts related to the current message and ${D_N}$ is the number of message response pairs in the training data. Our goal is to learn a dialogue model from $\mathcal{D}$ and $\mathcal{K}$. Given a new message $\bm{X}$, the model can identify related facts from $\mathcal{K}$ and generate response $\bm{Y}$.

\section{Model Framework of GenDS}
In this section, we will introduce the components of the GenDS system. 
The GenDS system has three components, which are listed as follows:
\begin{enumerate}
\item A candidate facts retriever first detects possible entities $\bm{E}$ in the input message $\bm{X}$, then retrieves a set of possible facts $\tau_{Q}$ from the knowledge base $\mathcal{K}$, based on the detected entities $\bm{E}$. 
\item A message encoder encodes the input message $\bm{X}$ into a set of intent vectors at each time step, denoted by $\bm{H}$.
\item A reply decoder takes $\bm{H}$ and $\tau_{Q}$ as input and generates the final response $\bm{Y}$ word by word. 
\end{enumerate}

\subsection{Candidate Facts Retriever}
The candidate facts retriever identifies facts that are related to the input message in the KB.
We denote the entities by $\bm{X}$ by $\bm{E} = (e_1,\cdots,e_m)$. $\bm{E}$ can be identified by keyword matching (e.g.,a singer, concert or song), or detected by more advanced methods such as entity linking or named entity recognition. 
Based on the detected entities $\bm{E}$, we can retrieve the relevant facts from $\mathcal{K}$. In traditional QA setting like GenQA, the assumption is that the subjects only appear in the input messages and the objects only appear in the responses. However, in the knowledge grounded conversation, the subject and object can occur together in one message or response. Thus, we retrieve facts with subjects matched with $\bm{E}$ and objectives matched with $\bm{E}$, denoted as $\tau_{QS} = \{\bm{\tau}_k\}_{k=1}^{K_{QS}}$ and $\tau_{QO} = \{\tau_k\}_{k=1}^{K_{QO}}$ respectively. We use $\tau_Q$ to denote the union of $\tau_{QS}$ and $\tau_{QO}$ as
$$ \tau_Q = \tau_{QS} \cup \tau_{QO}. $$
Notice that we do not restrict the amount of entities and retrieved facts, since different messages may need facts of variable sizes to generate reply. 

\subsection{Message Encoder}
The message encoder is designed to catch the user's intent. In this situation, the names of the entities are not essential and we can replace the entities by their types. For example, if we replace the entity in the message $<$recommend me songs of \textit{JAY} $>$ with its type, the transformed message $<$ recommend me songs of \textit{People} $>$ can still express the user's intent asking for song recommendation. After transformation, we do not need to learn the word embeddings of entities. Thus, we replace the entity in $\bm{X}$ by its type, and feed the transformed message into a RNN encoder word by word to get hidden representations $\bm{H}= \{\bm{h}_1,\cdots,\bm{h}_{M_T}\}$ of $\bm{X}$. 

\subsection{Reply Decoder}
The reply decoder generates the final response $\bm{Y}$ based on the user intention $\bm{H}$ and candidate facts $\tau_{Q}$. There are two categories of possible words in the generated response, the common words ($V_C$ ) and knowledge words ($V_E$)\footnote{Although some words may appear simultaneously in $V_C$ and $V_E$, they have different meaning. For example, "love" in $V_C$ is a verb, but may be one song name in $V_E$. Thus, we consider the word with same name in $V_C$ and $V_E$ as different word. In other words, there is no overlapping between $V_C$ and $V_E$. We also add the entity type and relations in $V_C$.}. We introduce a knowledge gate $z_t=\{0,1\}$ to determine which kind of words to be generated at each time step. In order to generate arbitrary number of possible entities in a single response, we propose a dynamic knowledge enquirer which can select entities according to the local contexts at various positions within a response. The dynamic knowledge enquirer can generate new knowledge words outside the scope of training data because it does not need to learn the embedding vectors for knowledge words. 

The probability of generating the answer $\bm{Y}=\{y_1,y_2,\cdots,y_{R_T}\}$ is defined as:
\begin{equation*}
\begin{split}
  & p(y_1,y_2,\cdots,y_{R_T} | \bm{H}, \tau_{Q}) \\
 &=\  p(y_1|\bm{H}, \tau_{Q}; \Theta) \prod_{t=2}^{R_T} p(y_t|y_1, ..., y_{t-1}, \bm{H}, \tau_{Q}; \Theta)
 \end{split}
\end{equation*}
where $\Theta$is all parameters in the GenDS model. The generation probability of $y_t$ is specified by
$$ p(y_t|y_1, ..., y_{t-1}, \bm{H}, \tau_{Q}; \Theta) = p(y_t| y_{t-1}, z_t, \bm{s}_t, \bm{H}, \tau_{Q}; \Theta) $$
where $\bm{s}_t$ is the hidden state of the decoder model and $z_t$ is the value of the knowledge gate $z_t$ at time step $t$. Based on the value of the knowledge gate $z_t$, the probability can be further decomposed as:
\begin{equation*}
\begin{split}
  &p(y_t| y_{t-1}, z_t, \bm{s}_t, \bm{H}, \tau_{Q}; \Theta)  \\
  &=\ p(y_t | z=0)\ p(z = 0 |y_{t-1}, \bm{s}_t, \bm{H}, \tau_{Q}; \Theta) \ +\\
  & \qquad p(y_t | z=1)\ p(z = 1|y_{t-1}, \bm{s}_t, \bm{H}, \tau_{Q}; \Theta) \\
  &=\  p_c( y_t )\ p(z = 0 | y_{t-1}, \bm{s}_t, \bm{H}, \tau_{Q}; \Theta) \ + \\
 & \qquad p_e( y_t )\ p(z = 1|y_{t-1}, \bm{s}_t, \bm{H}, \tau_{Q}; \Theta)
\end{split}
\end{equation*}
where $p_c(y_t)$ is the probability of $y_t$ generated by the common word generator and $p_e(y_t)$ is the probability of $y_t$ generated by the dynamic knowledge enquirer. 

\subsubsection{Common Word Generator}
The common word generator generates the common word $y_t$. Firstly, we calculate a message context vector $\bm{c}_t$ by using the attention mechanism~\cite{bahdanau2014neural} on the message hidden vectors $\bm{H}$ with the current generator hidden state $\bm{s}_{t-1}$ as:
$$ \bm{c}_t = \sum_{j=1}^T \alpha_{tj}\bm{h}_j. $$
where $\alpha_{tj}$ is computed by
$$ \alpha_{tj} = \frac{\exp(d_{tj})}{\sum_{k=1}^T \exp(d_{tk})};d_{tj} = {DNN}_1 (\bm{s}_{t-1}\bm{h}_j)$$
The generator hidden state is updated from the previous hidden state $\bm{s}_{t-1}$, the word embedding of previously predicted symbol $\mu_{y_{t-1}}$, and the message context vector $\bm{c}_t$. The hidden state of the common word generator is updated as:
$$ \bm{s}_t = \eta ({DNN}_1([\bm{s}_{t-1},\bm{c}_t]), \bm{\mu}_{y_{t-1}}) .$$
where $[.,.]$ denotes vector concatenation, and $\eta$ is a neural network function which is GRU \cite{bahdanau2014neural} in this paper. \\
The probability of generating common word $y_t$ is defined as:
$$ p_c(y_t) = {DNN}_1(\bm{s}_{t-1}) .$$
Since we do not learn the word embeddings of entities, we replace the entity by its  type, and use the entity type's word embedding instead. The entity type may discard some information for generating the following words. Thus, we fuse the word embedding of $y_{t-1}$ and the word embedding of $y_{t-2}$ into the word embedding of previous predicted symbols by one layer neural network.  

\subsubsection{Dynamic Knowledge Enquirer}
The dynamic knowledge enquirer generates knowledge word by ranking all entities in the retrieved facts $\tau_Q$ , according to their dynamic entity score. We regard objects in $\bm{\tau}_{QS}$ and subjects in $\bm{\tau}_{QO}$ as candidate entities. In order to generate multiple entities, the dynamic entity score incorporates both the message and the local context during decoding process. Specifically, we define three scores, the message matching score $\bm{r}_{e_k}$, the entity update score $f_t$ and the entity type update score $u_{kt}$, and the product of these scores as dynamic entity  score. 

The message matching score $\bm{r} \in \mathbb{R}^{|\tau_Q|}$ denotes the matching probability for each candidate entity $e_k$ in $\tau_Q$ with intent vectors of message.
The message matching score of candidate entity $e_k$ is obtained by a 2-layer neural network as follows:
\begin{equation}
{r_{e_k}=}
\begin{cases}
	{DNN}_2(\bm{h}_{M_T},\bm{\beta}_{e_k}), & for\ e_k \in \bm{e}_{Q}  \\
	0,                                                               & otherwise.
\end{cases}
\label{message matching score}
\end{equation}
where $\bm{h}_{M_T}$ is the last hidden state of message encoder, $\bm{E}_{\tau_Q}$ are entities in retrieved facts $\tau_{Q}$, and $\bm{\beta}_{e_k}$ is the concatenation of the word embedding of entity $e_k$'s type and corresponding predicate in retrieved fact. The message matching score is invariant during the decoding process. 

To take the history context into accounts, we compute the entity update score $\bm{f} \in \mathbb{R}^{|\tau_Q|}$ and the entity type update score $\bm{u} \in \mathbb{R}^{|\tau_Q|}$ respectively: 
\begin{eqnarray*}
\bm{f}_t &=& {DNN}_1(\bm{s}_t, \bm{\mu}_{y_{t-1}}, \bm{\mu}{y}_{t-1})\\
u_{kt} &=& {DNN}_1(\bm{s}_t, \bm{\mu}_{y_{t-1}}, \bm{\mu}_{e_k}) \quad for \ e_k \in \bm{E}. 
\end{eqnarray*}
where $\bm{y}_{t-1}$ is the one hot vector of last generated word $y_{t-1}$ and $\bm{\mu}_{e_k}$ is the word embedding of the entity type of $e_k$.

The entity update score is determined by the last generated word, and the entity type update score depends on the word embedding of last generated word and the entity's type embedding.

%Although the forgetting mechanism can decrease the probability of generated entities in future decoding, the generation probability of candidate entities is only determined by the question and entities' type. That means only one type can occur in the response. Besides, the question may not contain enough information to decide which entity occurs in the response. 

The final dynamic entity score $\bm{p}_{et}$ is computed as:
\begin{equation*}
p_e(y_t = et) = \\
\bm{p}_{et}  =  \bm{r} \circ \bm{f}_t \circ \bm{u}_t
\end{equation*}

\subsubsection{Final Response Generation with the knowledge gate}
In order to generate the final response with the common word generation and the dynamic knowledge enquirer, we introduce a binary knowledge gate $z_t \in \{0,1\}$ at each time step $t$. If $z_t$ equals $0$, the common word generate will be used to generate an common word $y_t$. If $z_t$ equals $1$, the dynamic knowledge enquirer will be used to generate an knowledge word. The knowledge gate is defined as 
$$p(z_t=1) = {DNN}_1(\bm{s}_t,\bm{c}_t,\bm{\mu}_{y_{t-1}})$$
where ${DNN}_1$ is one layer MLP,and $y_{t-1}$ will be replace with its type if it is entity.

In summary, the $y_t$ is generated as:
\begin{equation*}
\begin{split}
p(y_t|\bm{s}_t,y_{t-1},\bm{H}) 
 = & p(z =0|\bm{s}_t,y_{t-1},\bm{H})p_c(y_t|\bm{s}_t,y_{t-1},\bm{H}) \\
& +p(z = 1|\bm{s}_t,y_{t-1},\bm{H})\bm{p}_{et}
\end{split}
\end{equation*}

\subsection{Training}
For GenDS, We need to learn the parameters in message encoder, common word generator, and dynamic knowledge enquirer. In experiments, we found that if we strictly require the generated entities exactly same as ground truth, the model devotes to find entities same as ground truth and has little thinking of language model. This will degrade fluency of response. Thus, we train our system with multi-task learning \cite{zhang2017survey}:
\begin{enumerate}[1)]
\item  the model is trained with ground truth as output
\item  the output of task2 is to replace the entity in ground truth with its type
\end{enumerate}
We use our GenDS model for task 1, and the standard Seq2Seq model with attention \cite{bahdanau2014neural} for task 2.The task 2 can be regarded as the simplified version of task 1, whose goal is to generate fluent response and correct entity type. The task 2 can make up for the fluency ignorance of task 1. Two tasks share the message encoder, and common word generator decoder, and are trained with maximum likelihood estimation (MLE) as objective function.

% Four bar graphs
% \begin{figure}
% \centering
%     \subfigure[Music Convers - Accuracy]{
%     	\includegraphics[scale=0.25]{music_precision.pdf} 
%     }
% 	\subfigure[Music Convers - Recall]{
%     	\includegraphics[scale=0.25]{music_recall.pdf} 
% 	}
% 	\subfigure[Music QA - Accuracy]{
% 		\includegraphics[scale=0.25]{qa_precision.pdf} 
%     }
% 	\subfigure[Music QA - Recall]{ 
% 		\includegraphics[scale=0.25]{qa_recall.pdf} 
%     }
% 	\caption{The entity precision and recall for Music Convers and MusicQA}
% 	\label{entity-precision-recall}
% \end{figure}

\begin{figure*}[ht]
\centering
    \subfigure[MusicConvers - Accuracy]{
    	\includegraphics[scale=0.5]{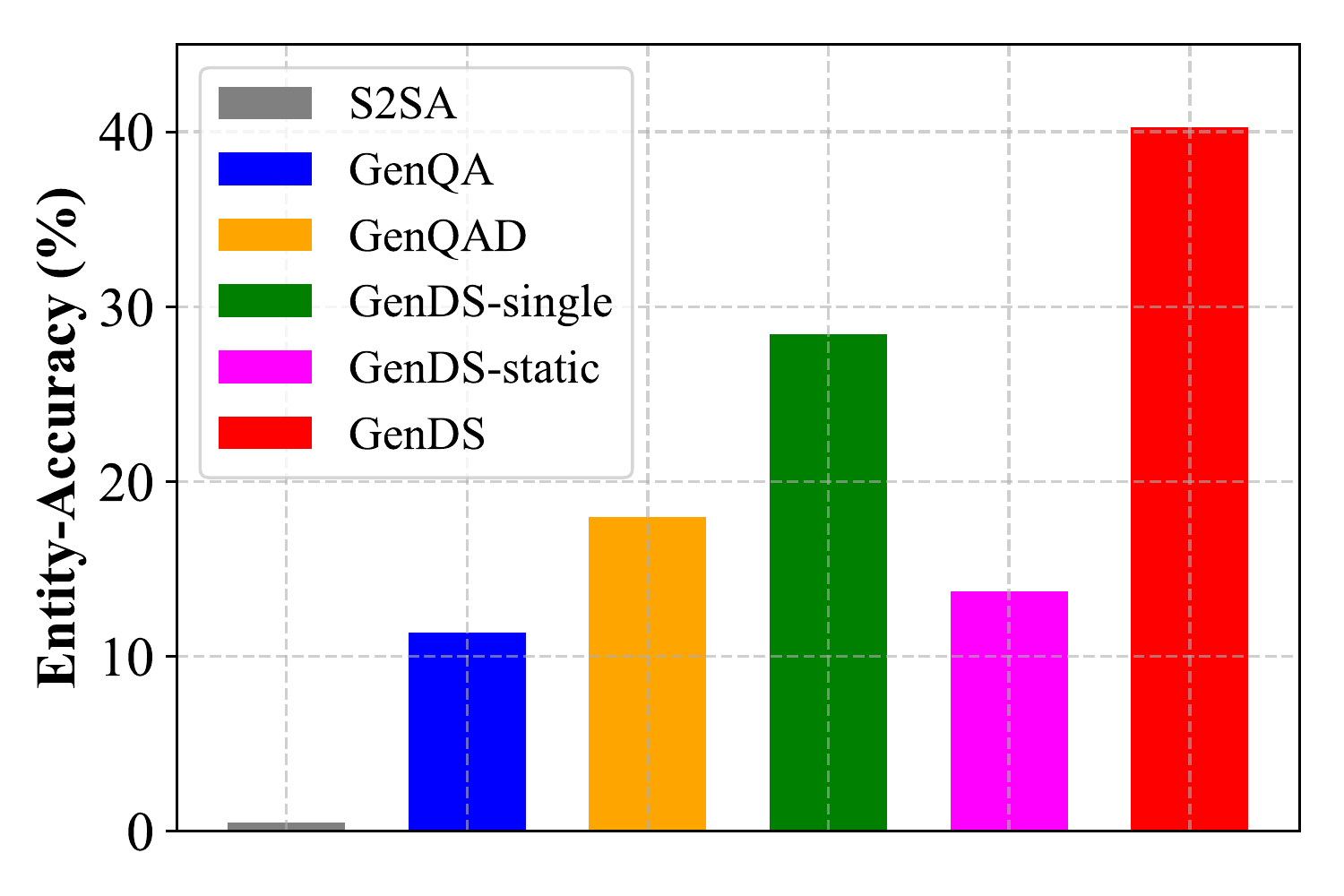} 
    }
	\subfigure[MusicConvers - Recall]{
    	\includegraphics[scale=0.5]{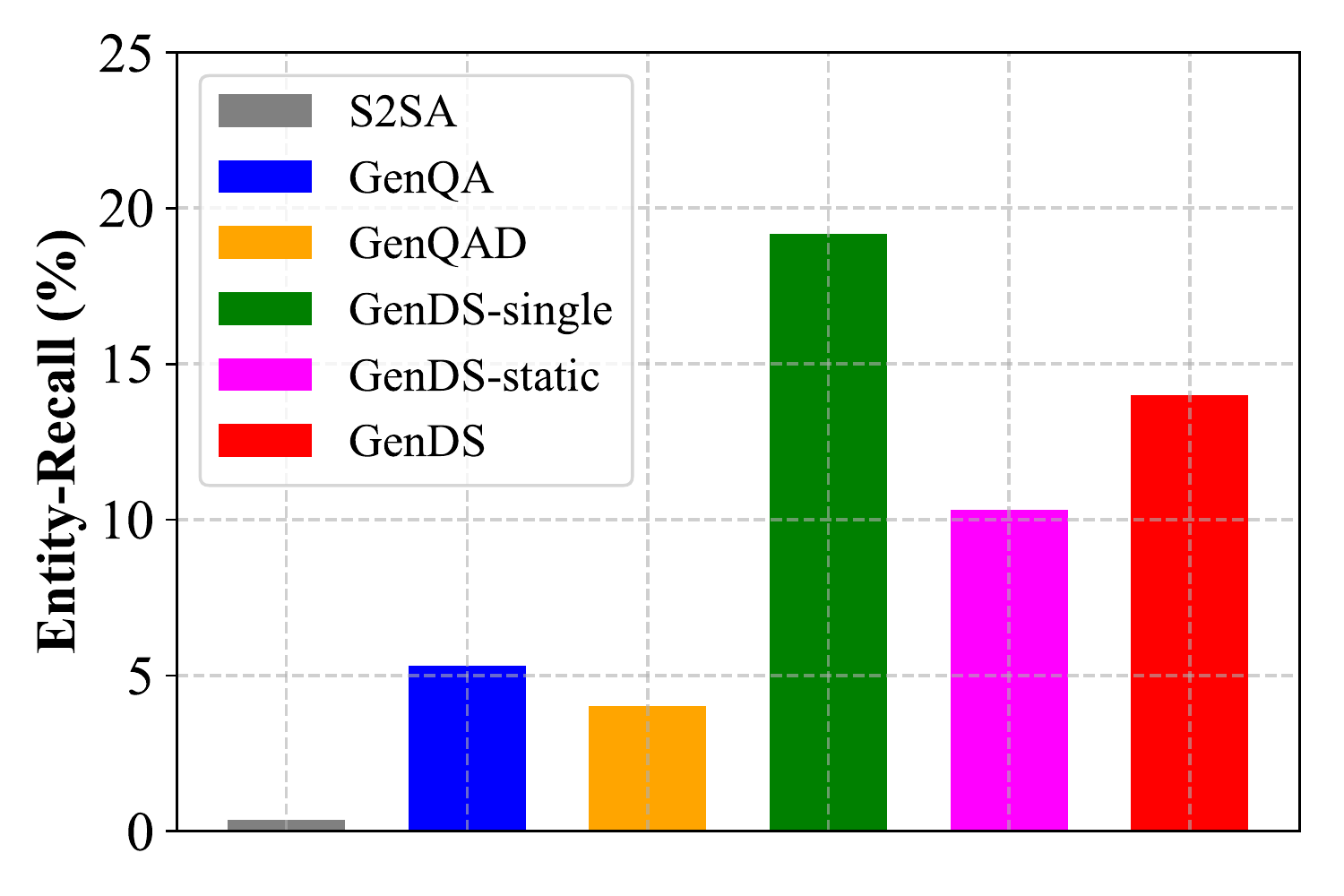} 
	}
	\caption{The entity precision and recall for MusicConvers }
	\label{MusicConvers-entity-precision-recall}
\end{figure*}
\begin{figure*}[ht]
\centering
	\subfigure[MusicQA - Accuracy]{
		\includegraphics[scale=0.5]{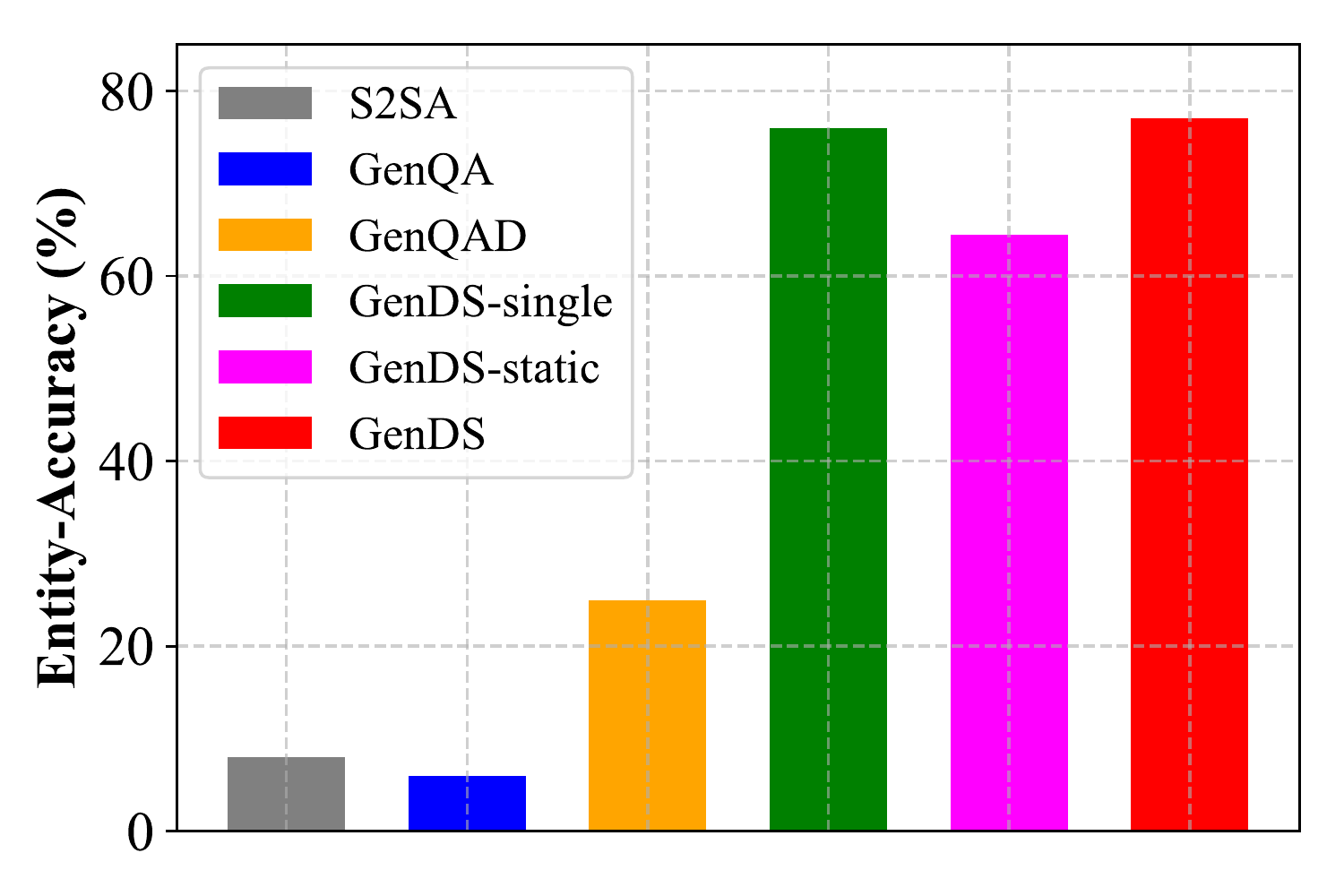} 
    }
	\subfigure[MusicQA - Recall]{ 
		\includegraphics[scale=0.5]{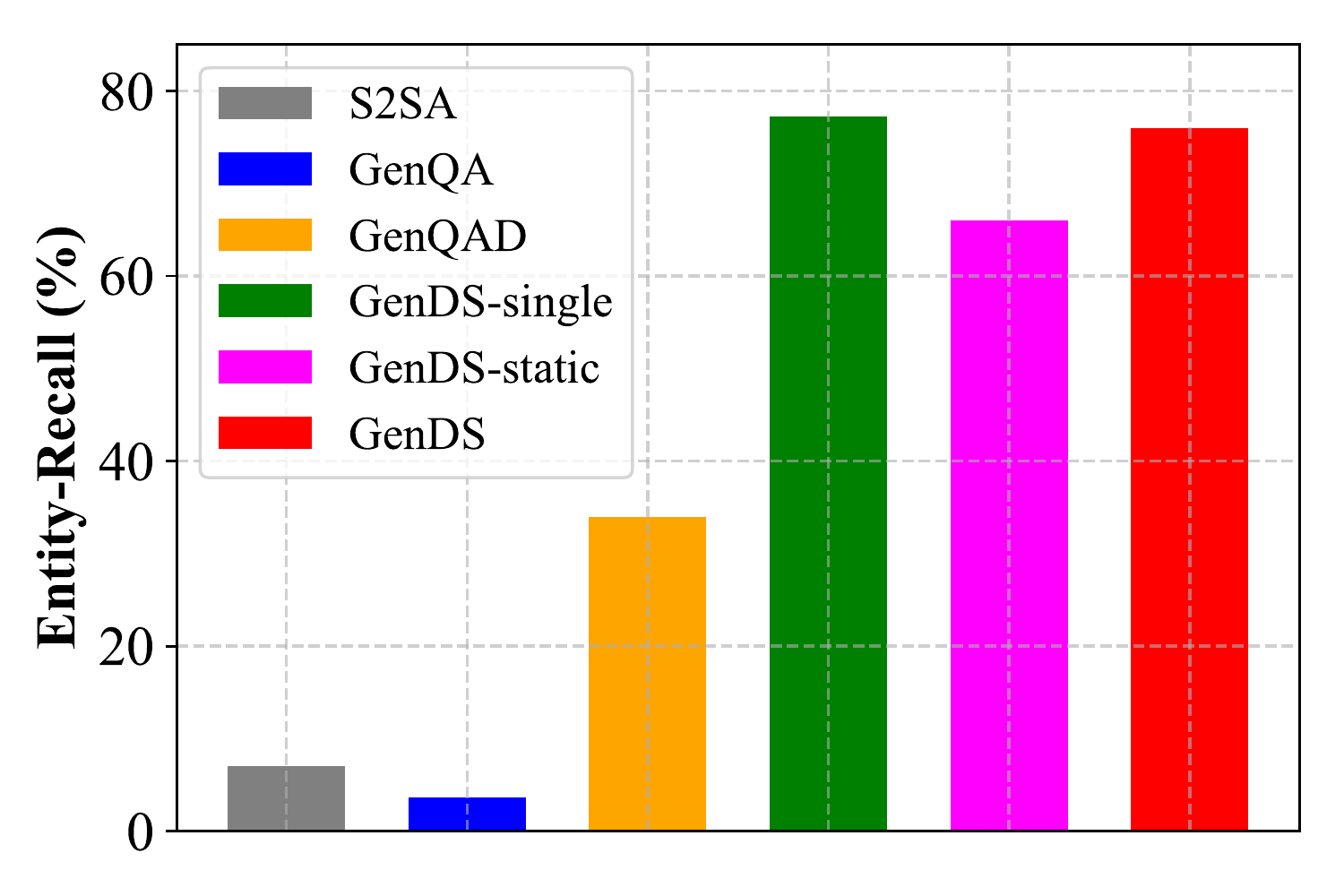} 
    }
	\caption{The entity precision and recall for MusicQA}
	\label{MusicQA-entity-precision-recall}
\end{figure*}

\section{Experiment}
In this section, we first describe the datasets used in the experiments. Then we describe the experiment setup and evaluation metrics. Finally, we present the experiment results in two datasets. 

\subsection{Dataset}
\subsubsection{MusicConvers}
We collect a human-to-human dialogue dataset from outsourcing in four months. All outsourcers are employed by an IT company in China. The new dataset named MusicConvers are composed of knowledge grounded conversations in music domain. Outsourcers  are asked to generate dialogues by talking about the music with their own knowledge. To simulate different individuals, speakers are given the predefined music knowledge as their private knowledge. Then two speakers start to talk based on given knowledge. We build the knowledge base by filtering the KB collected in \cite{yin2015neural}. The filtered KB is domain specific in music. Notice that the speakers may talk referring to their own KB. Hence, some new KB may appear in their conversations and is finally added to the given KB. To limit the range of topic, the given KB is restricted to one singer. We find that if speakers have different background knowledge, they are likely to talk more since they can obtain unseen knowledge from each other. Hence, the KBs shown to two speakers are overlapped but not exactly the same. We label the triples appear in each sentence. One example is shown in Figure \ref{Dialogue example}. Table \ref{tb:Music} shows the statistics of the dataset. We also find that the entities are sparse in our dataset. For example, $82\%$ entities occur less than three times. The entity sparsity conforms to common conversation style. In real life, individuals need only a small amount of common words to talk. But with the extension of their knowledge, they will enrich their vocabulary with more entities. 

\begin{table}[htbp]
\centering
\begin{tabular}{p{4.5cm}|p{1.0cm}} 
\hline
$\#$dialogues & 9993  \\ \hline  
vocabulary size  for message& 3256 \\ \hline
vocabulary size for response& 2976 \\ \hline
$\#$entities & 5988 \\ \hline 
$\#$ knowledge triples & 7612 \\ \hline
$\#$ relation in KB & 66 \\ \hline
\end{tabular} 
\caption{Statistics of the MusicConvers dataset}\label{tb:Music}
\end{table}

\subsubsection{Music Question Answering}
GenQA \cite{yin2015neural} is a large open-domain QA crawled from public websites, where each answer needs only one triple. However, in daily life, answers often consist of multiple facts. \cite{he2017hegenerating} extended the GenQA by adding more entities to questions. Unfortunately, this dataset has large redundancy. In other words the same question may appear several times. This redundancy may bring biases in the tests, since many questions may appear in the train data and test data simultaneously. Therefore, we remove those duplicated questions. We also filter out the QA pairs unrelated to music. The statistics of our music domain question answering are shown in Table \ref{tb:QA}. 

\begin{table}[htbp]
\centering
\begin{tabular}{p{4.5cm}|p{1.0cm}} 
\hline
$\#$QA pair & 30312  \\ \hline  
vocabulary size for message & 12576\\ \hline
vocabulary size for response & 13807 \\ \hline
$\#$ entities & 7176 \\ \hline 
$\#$ knowledge triples & 6238 \\ \hline
$\#$ relation in KB & 25 \\ \hline
\end{tabular} 
\caption{Statistics of the 	QA dataset}\label{tb:QA}
\end{table}

\subsection{Settings}
We adopt one-layer GRU \cite{cho2014properties} with 160 hidden units and 160-dimensional word vectors for both the message encoder and common word generator.  We use the Adam learning rule to update gradients in all experimental configurations. We train all models with learning rate as 1.0 for 5 epochs; after that, we decrease the learning rate by half and continue to train at least 5 epochs. Gradients are clipped at 5 to avoid gradient explosion. We randomly split the data into train ($80\%$) and test $20\%$). 

\subsection{Baselines}
We compare our model with seq2seq model with attention (\textbf{S2SA}), which is widely used in chit-chat dialogues system. To the best of our knowledge, there is no previous work on end-to-end knowledge grounded conversation with structured KB. Since existing generative QA models can be applied in knowledge grounded conversation, we also use generative QA model \textbf{GenQA} \cite{yin2015neural} as our baselines. To prove the effectiveness of dynamic knowledge enquirer, we improve GenQA with dynamic entity generation probability (\textbf{GenQAD}) , where the entity generation probability is determined by decoder hidden state $\bm{s}_t$, the intent vectors of message $\bm{H}$, and triple embedding $\bm{\mu}_{e_k}$. This is also the improvement of COREQA \cite{he2017hegenerating} for GenQA. We design the invariants of our model \textbf{GenDS-Single} and \textbf{GenDS-Static} to illustrate the benefits of multi-task and dynamic knowledge enquirer respectively. \textbf{GenDS-Single} is trained with single task where the output is the ground truth. \textbf{GenDS-Single} only uses the question matching score as the entity score Equation \ref{message matching score}, which is invariable during decoding.

\begin{table}[htp]
\centering
\begin{tabular}{|P{2.1cm}|P{0.8cm}|P{2cm}|P{2cm}|P{2cm}|} 
\hline
Models        &   BLEU      &        Precision    &       Recall       \\ \hline
S2SA         & 0.11        & $0.01 \pm 0.01$   & $0.004  \pm 0.02 $ \\ \hline
GenQA        & 0.05        & $0.1134 \pm 0.14$   & $0.05 \pm 0.1   $  \\ \hline
GenQAD       &  0.06       & $0.15   \pm 0.16$   & $0.05 \pm 0.1   $ \\ \hline 
GenDS-Single & 0.108       & $0.28 \pm 0.19$   & $\textbf{0.19}  \pm 0.18  $ \\ \hline
GenDS-Static & 0.108       & $0.14 \pm 0.15$   & $0.10 \pm 0.14  $ \\ \hline
GenDS        & \textbf{0.122} & $ \textbf{0.40 }\pm 0.25$   & $0.14   \pm 0.16  $ \\ \hline
\end{tabular} 
\caption{Automatic Evaluation on the Music dataset}
\label{tb:Music_Automatic_Result}
\end{table}

\begin{table}[htb]
\centering
\begin{tabular}{|P{2.2cm}|P{1.2cm}|P{1.6cm}|P{1.6cm}|} 
\hline
Models       & Grammar   & Context Relevance  &  Correctness   \\ \hline  
S2SA         &  1.76     & 0.87   & 0.16 \\ \hline
GenQA        &  1.28     & 0.95   & 0.41 \\ \hline
GenQAD       &  1.67     & 1.11   & 0.51 \\ \hline 
GenDS-Single &\textbf{2.16}& \textbf{1.67}   & \textbf{1.18} \\ \hline
GenDS-Static &  1.97     & 1.42   & 0.96 \\ \hline
GenDS        &  2.03     & 1.55   & 0.89 \\ \hline
\end{tabular} 
\caption{Human Evaluation on the MusicConvers dataset.}\label{tb:Music_Human_Result}
\end{table}

\begin{table}[htp]
\centering
\begin{tabular}{|P{2.1cm}|P{0.8cm}|P{2cm}|P{2cm}|P{2cm}|} 
\hline
Models        &    BLEU  &    Precision        &  Recall    \\ \hline
S2SA         & 0.05     & $0.08 \pm 0.125$   & $0.07 \pm 0.13  $ \\ \hline
GenQA        & 0.12     & $0.06 \pm 0.11$    & $0.04 \pm 0.09  $  \\ \hline
GenQAD       & 0.13     & $0.25 \pm 0.2 $    & $0.34 \pm 0.235 $ \\ \hline 
GenDS-Single & 0.226     & $0.76 \pm 0.205$   & $\textbf{0.77} \pm 0.21  $ \\ \hline
GenDS-Static & 0.19    & $0.64 \pm 0.23$    & $0.66 \pm 0.235 $ \\ \hline
GenDS        & \textbf{0.227}  & $ \textbf{0.77} \pm 0.205$   & $0.76   \pm 0.215  $ \\ \hline
\end{tabular} 
\caption{Automatic Evaluation on the QA dataset}
\label{tb:QA_Automatic_Result}
\end{table}

\subsection{Evaluation metrics}
We compare GenDS with baselines by both automatic evaluation and human evaluation. Following the existing works, we employ the BLEU \cite{papineni2002bleu} automatic evaluation. However, BLEU only reflects the words occurrence between the ground truth and the generated response. To measure the information correctness, we define the entity-accuracy, entity-recall inspired by the accuracy, recall in classification evaluation. The entity-accuracy is the percentage of right generated entities in generated entities, and the entity-recall is the percentage of right generated entities in ground truth entities. For entity-accuracy and entity-recall, we only handle the samples whose ground truths have entities. We also recruit human annotators to judge the quality of the generated responses with aspects of grammar, context relevance and information correctness. All scores range from $0$ to $3$. Higher score represents better performance in terms of the above three metrics.

\begin{table}[htp]
\centering
\begin{tabular}{|P{2cm}|P{5.5cm}|} 
\hline
\multirow{2}*{Message} & 
\small{I forgot what songs has \underline{Hu Ling} sung?}             \\
&
\small{\begin{CJK*}{UTF8}{gkai}忘记了，胡灵唱过什么歌？\end{CJK*}}  \\
\hline

\multirow{2}*{S2SA}  & 
\small{still \underline{New Experience}}       \\
&
\small{\begin{CJK*}{UTF8}{gkai}还有全新经验\end{CJK*}}  \\
\hline

\multirow{2}*{GenQA} & 
\small{\textbf{Meet in Guangzhou}, \textbf{Meet in Guangzhou}, \textbf{Meet in Guangzhou}, \textbf{Meet in Guangzhou} and so on}                         \\
&
\small{\begin{CJK*}{UTF8}{gkai}广州之约 ，广州之约 ，广州之约 ，广州之约 等等\end{CJK*}}   \\
\hline

\multirow{2}*{GenQAD} & 
\small{should be \textbf{I am still your lover}}                  \\
&
\small{\begin{CJK*}{UTF8}{gkai}就是 我依然是你的情人 \end{CJK*}}  \\
\hline

\multirow{2}*{GenDS-Static} & 
\small{\textbf{Song of Wind}, \textbf{Song of Wind}}                 \\
&
\small{\begin{CJK*}{UTF8}{gkai}听风的歌 ，听风的歌\end{CJK*}}    \\
\hline

\multirow{2}*{GenDS-Single}  &
\small{\textbf{I am still your lover}, \textbf{Song of Wind}}                        \\
&
\small{\begin{CJK*}{UTF8}{gkai}我依然是你的情人，听风的歌 \end{CJK*} }           \\
\hline
\end{tabular} 
\caption{Case Study For Knowledge Grounded Conversation. Entities are underlined, and true entities are in bold.}\label{tb:Case_Conversation_for_KGC}
\end{table}

\begin{table}[htp]
\centering
\begin{tabular}{|P{2cm}|P{5.5cm} |} 
\hline

 \multirow{2}*{Message} &
 \small{Didn't know this singer, do you know \underline{Zhiqian Xue}} \\
  &
 \small{\begin{CJK*}{UTF8}{gkai}没听过这个歌手哎，你知道薛之谦么\end{CJK*}}  \\
 \hline
 
 \multirow{2}*{S2SA} &
 \small{I knew her songs} \\
  &
 \small{\begin{CJK*}{UTF8}{gkai}听过她的歌呀\end{CJK*}}  \\
 \hline
 
 \multirow{2}*{GenQA} &
  \small{Animal World singing the song is that he wrote the lyrics of the day more than a few days} \\ 
  &
  \small{\begin{CJK*}{UTF8}{gkai}动物世界 唱的歌 都 是 他 写 填词 日 超多 日 超多 日\end{CJK*}}  \\
  \hline
  
 \multirow{2}*{GenQAD} &
  \small{Heard of these two} \\
  &
 \small{\begin{CJK*}{UTF8}{gkai}听 过 这两首 是 听 过 这两首 了 哦\end{CJK*}}   \\
 \hline
 
 \multirow{2}*{GenDS-Single} &
  \small{\underline{\textbf{Prince Back}} is the song of \underline{\textbf{Zhiqian Xue}}, all written by himself} \\
&
 \small{\begin{CJK*}{UTF8}{gkai}王子归来 是 薛之谦 的 歌 ， 都 是 作词 。\end{CJK*} } \\
\hline
\end{tabular} 
\caption{Case Study For Unseen Knowledge Words. Unseen Entities are underlined, and true entities are in bold.}  \label{tb:Case_Conversation_for_UKW}
\end{table}

\subsection{Evaluation Results}
\textbf{MusicConvers:} The BLEU score computed with 1-ngram is shown in Table \ref{tb:Music_Automatic_Result}. Figure \ref{MusicConvers-entity-precision-recall} shows the entity-accuracy and entity-recall. The human evaluation result is shown in Table \ref{tb:Music_Human_Result}. We compute the mean score of each metric. For automatic evaluation, GenDS shows the best performance on BLEU and entity-accuracy, while GenDS-single achieves the highest entity-recall. Although GenDS does not overwhelm on S2SA in terms of BLEU, it improves entity accuracy and recall by $39\%$ and $14\%$ respectively. This indicates that GenDS can reply the message with more correct information. S2SA cannot respond with correct information, which is mainly due to the lack of grounding into external knowledge. Although GenQA and GenQAD can incorporate KB in response, their performance on entity-accuracy and entity-recall still cannot compete with GenDS.  For GenQA, the entity generation probability is fixed during decoding. As a consequence,  GenQA cannot generate different entities. For GenQAD, its update mechanism of entity generation probability is less effective than our dynamic knowledge enquirer. BLEU scores of GenQA and GenQAD are lowest among all models. We think that MusicConvers has no enough data for GenQA and GenQAD  to learn reliable entity representations.  This illustrate thet GenDS can achieve decent performance even on small dataset. GenDS achieves higher BLEU than GenDS-Single, which confirms the benefit of multi-task learning for improving the fluency. For human evaluation, the GenDS-single achieves the best performance in terms of grammar, context relevance and information correctness. Although S2SA can generate fluent responses, these responses contain little correct information and are less semantically relevant with the message. The performance of GenDS is slightly worse than GenDS-single. We infer that is due to the task 2, where GenDS tends to generate fluent response instead of correct information. 

\textbf{MusicQA} The automatic evaluation results on MusicQA are shown in Table \ref{tb:QA_Automatic_Result} and Figure \ref{MusicQA-entity-precision-recall}. GenDS improves BLEU score, entity-accuracy and entity-recall significantly compared with S2SA, GenQA and GenQAD. GenQA does not obtain comparable performance with the original QA dataset \cite{yin2015neural}. This may be the due to our mitigation of redundancy for the dataset. Unlike MusicConvers, GenDS-static exhibits decent performance on entity-accuracy and entity-recall. $99\%$ questions in MusicQA only contain one entity in answer. Thus, most of messages do not need dynamic knowledge enquirer to generate multiple entities. However, GenDS still achieves higher entity-accuracy and entity-recall than GenDS-static. This verifies that the proposed dynamic knowledge enquirer is useful even when only one entity is generated.

\subsection{Case Studies}

Table \ref{tb:Case_Conversation_for_KGC} compares models with some examples in text data. Entities are underlined, and true generated entities are in bold. Although S2SA may generate response with entities, it hardly generates true entities. Without dynamic knowledge word generation probability, GenDS-static and GenQA can not generate different entities in one response. GenDS-Single can generate multiple entities. This indicates that the dynamic knowledge enquirer learns to punish the generated words. To verify the validity on unseen entities, we expand the KB with new knowledge triples, and outsourcers provide the input messages based on the new KB. Table \ref{tb:Case_Conversation_for_UKW} shows the responses of these messages, where unseen entities are underlines, and true generated entities are in bold. GenDS-single can generate the decent response with multiple correct entities even when the entities in the input message are not included in training data. Besides, GenDS-single can use the unseen entities in new KB as response.  Table \ref{tb:Case_Conversation_for_UKW} shows that ability of GenDS-Single to generate different entity types in one response. In this example, the singer is generated after the song.  Such ability indicates that the dynamic knowledge enquirer can find the co-occurrence of different entities to entities with different types.

\subsection{Conclusion}
We propose an end-to-end knowledge grounded conversation model, GenDS, to incorporate structured KB in response generation. The model can generate responses with any number of answer entities, even when these entities never appear in the training set. It outperforms traditional non-goal-driven dialogue system S2SA and generative QA models on MusicConvers and MusicQA datasets. Being able to deal with unseen entities, GenDS is scalable with new KB. For further work, we plan to improve the GenDS with transfer learning \cite{pan2010survey}, such that GenDS can be transferred to another domain like sport. 

\bibliography{reference}

\begin{thebibliography}{}

\bibitem[\protect\citeauthoryear{Bahdanau, Cho, and
  Bengio}{2014}]{bahdanau2014neural}
Bahdanau, D.; Cho, K.; and Bengio, Y.
\newblock 2014.
\newblock Neural machine translation by jointly learning to align and
  translate.
\newblock {\em arXiv preprint arXiv:1409.0473}.

\bibitem[\protect\citeauthoryear{Cho \bgroup et al\mbox.\egroup
  }{2014}]{cho2014properties}
Cho, K.; Van~Merri{\"e}nboer, B.; Bahdanau, D.; and Bengio, Y.
\newblock 2014.
\newblock On the properties of neural machine translation: Encoder-decoder
  approaches.
\newblock {\em arXiv preprint arXiv:1409.1259}.

\bibitem[\protect\citeauthoryear{Ghazvininejad \bgroup et al\mbox.\egroup
  }{2017}]{ghazvininejad2017knowledge}
Ghazvininejad, M.; Brockett, C.; Chang, M.-W.; Dolan, B.; Gao, J.; Yih, W.-t.;
  and Galley, M.
\newblock 2017.
\newblock A knowledge-grounded neural conversation model.
\newblock {\em arXiv preprint arXiv:1702.01932}.

\bibitem[\protect\citeauthoryear{Han \bgroup et al\mbox.\egroup
  }{2015}]{han2015exploiting}
Han, S.; Bang, J.; Ryu, S.; and Lee, G.~G.
\newblock 2015.
\newblock Exploiting knowledge base to generate responses for natural language
  dialog listening agents.
\newblock In {\em SIGDIAL Conference},  129--133.

\bibitem[\protect\citeauthoryear{Hao \bgroup et al\mbox.\egroup
  }{2017}]{haoend2017QA}
Hao, Y.; Zhang, Y.; Liu, K.; He, S.; Liu, Z.; Wu, H.; and Zhao, J.
\newblock 2017.
\newblock An end-to-end model for question answering over knowledge base with
  cross-attention combining global knowledge.
\newblock In {\em Proceedings of the 40th annual meeting on association for
  computational linguistics},  3094--3100.

\bibitem[\protect\citeauthoryear{He \bgroup et al\mbox.\egroup
  }{2017}]{he2017hegenerating}
He, S.; Liu, C.; Liu, K.; and Zhao, J.
\newblock 2017.
\newblock Generating natural answers by incorporating copying and retrieving
  mechanisms in sequence-to-sequence learning.
\newblock In {\em ACL}.

\bibitem[\protect\citeauthoryear{Pan and Yang}{2010}]{pan2010survey}
Pan, S.~J., and Yang, Q.
\newblock 2010.
\newblock A survey on transfer learning.
\newblock {\em IEEE Transactions on knowledge and data engineering}
  22(10):1345--1359.

\bibitem[\protect\citeauthoryear{Papineni \bgroup et al\mbox.\egroup
  }{2002}]{papineni2002bleu}
Papineni, K.; Roukos, S.; Ward, T.; and Zhu, W.-J.
\newblock 2002.
\newblock Bleu: a method for automatic evaluation of machine translation.
\newblock In {\em Proceedings of the 40th annual meeting on association for
  computational linguistics},  311--318.
\newblock Association for Computational Linguistics.

\bibitem[\protect\citeauthoryear{Serban \bgroup et al\mbox.\egroup
  }{2016}]{serban2016building}
Serban, I.~V.; Sordoni, A.; Bengio, Y.; Courville, A.~C.; and Pineau, J.
\newblock 2016.
\newblock Building end-to-end dialogue systems using generative hierarchical
  neural network models.
\newblock In {\em AAAI},  3776--3784.

\bibitem[\protect\citeauthoryear{Serban \bgroup et al\mbox.\egroup
  }{2017}]{serban2017hierarchical}
Serban, I.~V.; Sordoni, A.; Lowe, R.; Charlin, L.; Pineau, J.; Courville,
  A.~C.; and Bengio, Y.
\newblock 2017.
\newblock A hierarchical latent variable encoder-decoder model for generating
  dialogues.
\newblock In {\em AAAI},  3295--3301.

\bibitem[\protect\citeauthoryear{Shang, Lu, and Li}{2015}]{shang2015neural}
Shang, L.; Lu, Z.; and Li, H.
\newblock 2015.
\newblock Neural responding machine for short-text conversation.
\newblock {\em arXiv preprint arXiv:1503.02364}.

\bibitem[\protect\citeauthoryear{Sordoni \bgroup et al\mbox.\egroup
  }{2015}]{sordoni2015neural}
Sordoni, A.; Galley, M.; Auli, M.; Brockett, C.; Ji, Y.; Mitchell, M.; Nie,
  J.-Y.; Gao, J.; and Dolan, B.
\newblock 2015.
\newblock A neural network approach to context-sensitive generation of
  conversational responses.
\newblock {\em arXiv preprint arXiv:1506.06714}.

\bibitem[\protect\citeauthoryear{Sutskever, Vinyals, and
  Le}{2014}]{sutskever2014sequence}
Sutskever, I.; Vinyals, O.; and Le, Q.~V.
\newblock 2014.
\newblock Sequence to sequence learning with neural networks.
\newblock In {\em Advances in neural information processing systems},
  3104--3112.

\bibitem[\protect\citeauthoryear{Yao, Zweig, and Peng}{2015}]{yao2015attention}
Yao, K.; Zweig, G.; and Peng, B.
\newblock 2015.
\newblock Attention with intention for a neural network conversation model.
\newblock {\em arXiv preprint arXiv:1510.08565}.

\bibitem[\protect\citeauthoryear{Yin \bgroup et al\mbox.\egroup
  }{2015}]{yin2015neural}
Yin, J.; Jiang, X.; Lu, Z.; Shang, L.; Li, H.; and Li, X.
\newblock 2015.
\newblock Neural generative question answering.
\newblock {\em arXiv preprint arXiv:1512.01337}.

\bibitem[\protect\citeauthoryear{Zhang and Yang}{2017}]{zhang2017survey}
Zhang, Y., and Yang, Q.
\newblock 2017.
\newblock A survey on multi-task learning.
\newblock {\em arXiv preprint arXiv:1707.08114}.

\end{thebibliography}
\bibliographystyle{aaai}

\end{document}